\documentclass{article}
\usepackage{spconf,amsmath,graphicx,hyperref}
\usepackage{xcolor}

\usepackage{amssymb}
\usepackage{booktabs}
\usepackage{multicol}
\usepackage{multirow}
\usepackage{graphicx}
\usepackage[table]{xcolor}
\usepackage{caption}

\title{Latent Temporal Discrepancy as Motion Prior: \\ A Loss-Weighting Strategy for Dynamic Fidelity in T2V}

\name{
\begin{tabular}{c}
Meiqi Wu$^{1,4}$\sthanks{Equal contribution}\sthanks{Work done during the internship at AMAP, Alibaba Group.},
    Bingze Song$^{2}$\footnotemark[1],
    Ruimin Lin$^{2}$,
    Chen Zhu$^{3}$\footnotemark[2],
    Xiaokun Feng$^{1,4}$\footnotemark[2],\\
    Jiahong Wu$^{2}$\sthanks{Corresponding author.},
    Xiangxiang Chu$^{2}$,
    Kaiqi Huang$^{1,4}$\footnotemark[3]
\end{tabular}}
\address{$^1$ University of Chinese Academy of Sciences, Beijing, China \\
        $^2$ AMAP, Alibaba Group, Beijing, China \quad $^3$ Southeast University, Nanjing, China   \\ 
        $^4$ CRISE, Institute of Automation Chinese Academy of Sciences, Beijing, China}

\begin{document}

\maketitle

\begin{abstract}
Video generation models have achieved notable progress in static scenarios, yet their performance in motion video generation remains limited, with quality degrading under drastic dynamic changes. This is due to noise disrupting temporal coherence and increasing the difficulty of learning dynamic regions. 
{Unfortunately, existing diffusion models rely on static loss for all scenarios, constraining their ability to capture complex dynamics.}
To address this issue, we introduce Latent Temporal Discrepancy (LTD) as a motion prior to guide loss weighting. LTD measures frame-to-frame variation in the latent space, assigning larger penalties to regions with higher discrepancy while maintaining regular optimization for stable regions. This motion-aware strategy stabilizes training and enables the model to better reconstruct high-frequency dynamics. Extensive experiments on the general benchmark VBench and the motion-focused VMBench show consistent gains, with our method outperforming strong baselines by 3.31\% on VBench and 3.58\% on VMBench, achieving significant improvements in motion quality. 
\end{abstract}


\vspace{-3pt}
\section{Introduction}
\vspace{-3pt}
Video generation models have achieved remarkable progress~\cite{wang2023modelscope,yang2024cogvideox,opensora2,Sora,wan2.1}, particularly in low-motion scenarios. However, performance degrades on segments with drastic dynamic changes ($e.g.
$, {high-frequency motion, drastic dynamics, fluid simulation}), with training losses exhibiting instability. 
While motion priors are introduced in video action generation to address its inherent difficulties, the core challenge lies in the precise acquisition and efficient computation of these priors. 

Previous methods primarily calculate dynamic scores directly from the video itself to quantify motion priors. 
Mask-based methods \cite{dai2023animateanything, ma2024follow, yariv2025through} utilize masks to delineate regions of motion, thereby identifying motion areas. However, the acquisition of motion regions and the scope of mask movement are limited, often exclusively optimizing for the inference stage. 
Optical flow-based methods \cite{ma2024follow, wang2025motif, nam2025optical} for motion prior computation offer a more refined capture of actions. Nevertheless, these methods are characterized by high computational cost, complexity, and inherent limitations in capturing non-displacement dynamics. 
Methods based on inter-frame differences \cite{chen2024livephoto, ma2024cinemo} also represent a viable approach, such as frame residuals and structural similarity index.

\begin{figure}
    \centering
    \includegraphics[width=\linewidth]{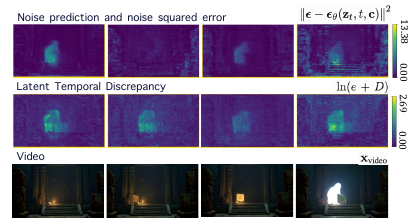}
    \captionsetup{aboveskip=2pt, belowskip=0pt}
    \caption{Noise prediction and noise squared error $\left\| \boldsymbol{\epsilon} - \boldsymbol{\epsilon}_\theta(\mathbf{z}_t, t, \mathbf{c}) \right\|^2 $, Latent Temporal Discrepancy $\mathrm{ln}(e+D)$, and Video $\mathbf{x}_\mathrm{video}$ visualization.}
    \vspace{-3pt}
    \label{fig:method-f}
\end{figure}

To address this, we instead derive a motion prior directly from the model’s internal representations. Concretely, we compute a \textbf{L}atent \textbf{T}emporal \textbf{D}iscrepancy (LTD) between adjacent frames and use it to guide loss reweighting toward highly dynamic regions while retaining regular optimization elsewhere (Fig.~\ref{fig:method-f}). This LTD-guided loss stabilizes training and encourages the capture of high-frequency dynamics without external motion estimation.

\begin{figure*}
    \centering
    \includegraphics[width=\linewidth]{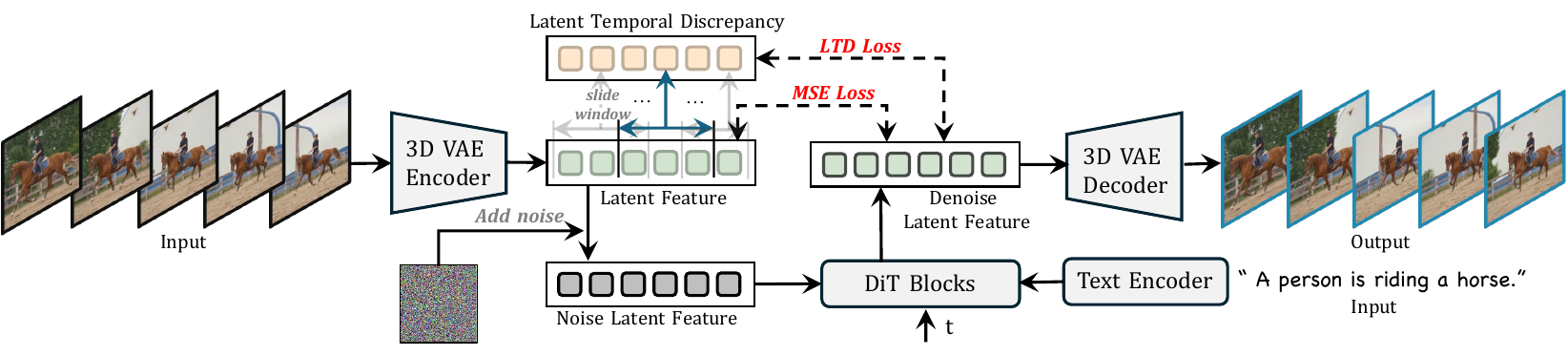}
    \captionsetup{aboveskip=2pt, belowskip=0pt}
    \caption{Overview of our method. We use a 3D VAE for video encoding and decoding, and adopt DiT as the diffusion model. A motion prior is built from temporal discrepancies of latent features in a sliding window, which guides the LTD Loss to enhance learning of motion.}
    \vspace{-3pt}
    \label{fig:method}
\end{figure*}

Across extensive experiments on VBench~\cite{huang2024vbench} and the motion-focused VMBench~\cite{ling2025vmbench}, our method consistently outperforms a strong baseline (Wan2.1~\cite{wan2.1}), achieving improvements of 3.31\% and 3.58\%, respectively. Moreover, when compared against other T2V —VideoCrafter~\cite{chen2024videocrafter2}, CogVideoX~\cite{yang2024cogvideox}, Hunyuan~\cite{kong2024hunyuanvideo}, OpenSora~\cite{opensora2}, OpenSora-plan~\cite{lin2024opensora-plan}—our approach remains highly competitive and often superior. The gains concentrate on dynamic metrics—temporal consistency, motion smoothness, and large-motion fidelity—while maintaining parity on static quality ($e.g.$, appearance and text alignment). In addition, the LTD-guided loss effectively reduces training-loss instability in high-motion segments.

\textbf{Contributions}: (1) A flow-free motion prior, \textbf{LTD}, that responds to both displacement and non-displacement dynamics. (2) A simple LTD-guided loss weighting scheme that improves stability and dynamic fidelity and is plug-and-play for latent diffusion. (3) Extensive experiments on VBench and the motion-focused VMBench, showing consistent gains ($e.g.$, 3.31\% on VBench and 3.58\% on VMBench) over strong baselines.

\vspace{-3pt}
\section{Method}
\vspace{-3pt}

The T2V model aims to generate an $L$-frame video ${\mathbf{x}}=\{x_0,x_1,...,x_L\}$ from a text caption. 
However, generating high-quality motion has long been a challenging aspect.
Moreover, during training, video segments with higher motion intensity often exhibit greater losses, as shown in Fig.\ref{fig:method-s}.
To address this, this section proposes a method based on latent temporal discrepancy to extract motion priors, thereby applying stronger loss weights to video segments with high motion intensity and enhancing the model's ability to generate dynamic content.

\vspace{-3pt}
\subsection{Preliminaries}
\vspace{-3pt}
\textbf{Latent Diffusion Models} \cite{rombach2022high} are a class of generative models that adapt the principles of Denoising Diffusion Probabilistic Models \cite{ho2020denoising, song2020score} to operate within a compressed latent space, thereby achieving significant gains in computational efficiency while maintaining high generation quality. The architecture first employs a pretrained autoencoder, consisting of an encoder $\mathcal{E}$ and a decoder $\mathcal{D}$, to learn a latent space. An input video $\mathbf{x}_{\text{video}} \in \mathbb{R}^{F \times H \times W \times C}$ is mapped by the encoder to an initial latent representation $\mathbf{z}_0 = \mathcal{E}(\mathbf{x}_{\text{video}}), \mathbf{z}_0 \in \mathbb{R}^{F_l \times H_l \times W_l \times C_l}$, compressing the data before the diffusion process begins.

The forward diffusion process progressively adds Gaussian noise to the initial latent $\mathbf{z}_0$ over $T$ timesteps. The noisy latent at any step $t$, $\mathbf{z}_t$, can be sampled in a closed form: $\mathbf{z}_t = \sqrt{\bar{\alpha}_t}\mathbf{z}_0 + \sqrt{1 - \bar{\alpha}_t}\boldsymbol{\epsilon}$. Here, $\boldsymbol{\epsilon}$ is standard Gaussian noise, and $\bar{\alpha}_t$ is a coefficient from a predefined noise schedule. At the end of the process, $\mathbf{z}_T$ approximates an isotropic Gaussian distribution.

The learning objective is to train a neural network  $\boldsymbol{\epsilon}_\theta$, to reverse this noising process. This constitutes reverse denoising process. The network is tasked with predicting the noise component $\boldsymbol{\epsilon}$ from the noisy latent $\mathbf{z}_t$, conditioned on the timestep $t$ and optional external context $\mathbf{c}$ (such as text embeddings). The model is trained by minimizing a simplified objective function, which is a mean squared error between the true and predicted noise:

{\small
\begin{equation}
    \mathcal{L}_{\text{diffusion}} = \mathbb{E}_{\mathbf{z}_0, \boldsymbol{\epsilon}, t, \mathbf{c}} \left[ \left\| \boldsymbol{\epsilon} - \boldsymbol{\epsilon}_\theta(\mathbf{z}_t, t, \mathbf{c}) \right\|^2 \right].
\end{equation}}

During inference, a new video is generated by first sampling a random tensor from the prior distribution, $\mathbf{z}_T \sim \mathcal{N}(\mathbf{0}, \mathbf{I})$, and then iteratively applying the learned denoising network $\boldsymbol{\epsilon}_\theta$ to reverse the diffusion process from $t=T$ down to $t=1$, eventually yielding a clean latent sample $\mathbf{z}_0$. Finally, the pretrained decoder $\mathcal{D}$ maps this latent representation back into the pixel space to synthesize the final video output.

\vspace{-3pt}
\subsection{Modeling}
\vspace{-3pt}

Although $\mathcal{L}_\mathrm{diffusion}$ can generate usable videos to some extent during the diffusion process, it is often insufficient to ensure high generation quality for modalities with very high information density, such as videos. In particular, in motion regions, this can easily lead to motion distortion or unrealistic behavior. 
To improve motion modeling, we incorporate a motion prior into diffusion loss through a reweighting mechanism for high-dynamic regions, inspired by the principle of Focal Loss \cite{lin2017focal} that emphasizes challenging motion patterns, enhances their gradient contributions, and promotes more accurate motion generation.

\begin{table*}[t]
\caption{Comparison to prior works on VBench~\cite{zheng2025vbench}. The top half shows results on VBench prompts, and the bottom half on the motion-oriented VMBench prompts.}
\label{tb:sota}
\centering
\resizebox{\linewidth}{!}{
\begin{tabular}{l|cccccccc}
\toprule
Models &
\begin{tabular}[c]{@{}c@{}}Aesthetic\\ Quality(\%) $\uparrow$\end{tabular} &
\begin{tabular}[c]{@{}c@{}}Background\\ Consistency(\%) $\uparrow$\end{tabular} &
\begin{tabular}[c]{@{}c@{}}Dynamic\\ Degree(\%) $\uparrow$\end{tabular} &
\begin{tabular}[c]{@{}c@{}}Image\\ Quality(\%) $\uparrow$\end{tabular} &
\begin{tabular}[c]{@{}c@{}}Motion\\ Smoothness(\%) $\uparrow$\end{tabular} &
\begin{tabular}[c]{@{}c@{}}Subject\\ Consistency(\%) $\uparrow$\end{tabular} &
\begin{tabular}[c]{@{}c@{}}Temporal\\ Flickering(\%) $\uparrow$ \end{tabular}&
\begin{tabular}[c]{@{}c@{}}Quality\\ Score(\%) $\uparrow$ \end{tabular} 
 \\
\hline
\rowcolor{gray!20}
&\multicolumn{8}{c}{\textit{\textbf{VBench-prompt}}}\\
VideoCrafter & 63.13 & 98.22 & 42.50 & 67.22 & 97.73 & 96.85 & 98.41 & 82.20 \\
Wan2.1 (Base)~\cite{wan2.1}            & 64.41            & 96.26           & {70.83}   & 65.06             & 97.88      & 93.49        & 99.51       & 83.76  \\
CogVideoX~\cite{yang2024cogvideox} & 60.82 & 96.63 & 59.86 & 61.68 & 97.73 & 96.78 & 98.89 & 81.43 \\
Hunyuan~\cite{kong2024hunyuanvideo} & 65.24 & 93.08 & \textbf{87.50} & 61.59 & 98.34 & 88.60 & 98.38 & 82.87 \\
OpenSora~\cite{opensora2}   & 64.33 & \textbf{98.00} & 20.74 & 65.62 & \textbf{99.49} & \textbf{98.75} & 99.40 & 80.14 \\
OpenSora-Plan & 56.85 & 96.73 & 47.72 & 62.28 & 98.28 & 95.73 & 99.03 & 80.91 \\
\textbf{Ours}                   & \textbf{65.80}   & {97.16}  & 68.06            & \textbf{67.85}    & {98.39}   & 95.65   & \textbf{99.69} & \textbf{85.11} \\ 
\hline
\rowcolor{gray!20}

&\multicolumn{8}{c}{\textit{\textbf{VMBench-prompt}}}\\
Wan2.1 (Base)~\cite{wan2.1} & \textbf{65.02} & 91.57 & 33.33 & 62.85 & 97.50 & 90.11 & 73.40 & 77.27 \\
\textbf{Ours} & 55.25 & \textbf{92.59} & \textbf{78.10} & \textbf{66.40} & \textbf{97.93} & \textbf{91.90} & \textbf{80.36} & \textbf{80.85} \\
\bottomrule
\end{tabular}
}

\end{table*}

\begin{figure}
    \centering
    \includegraphics[width=\linewidth]{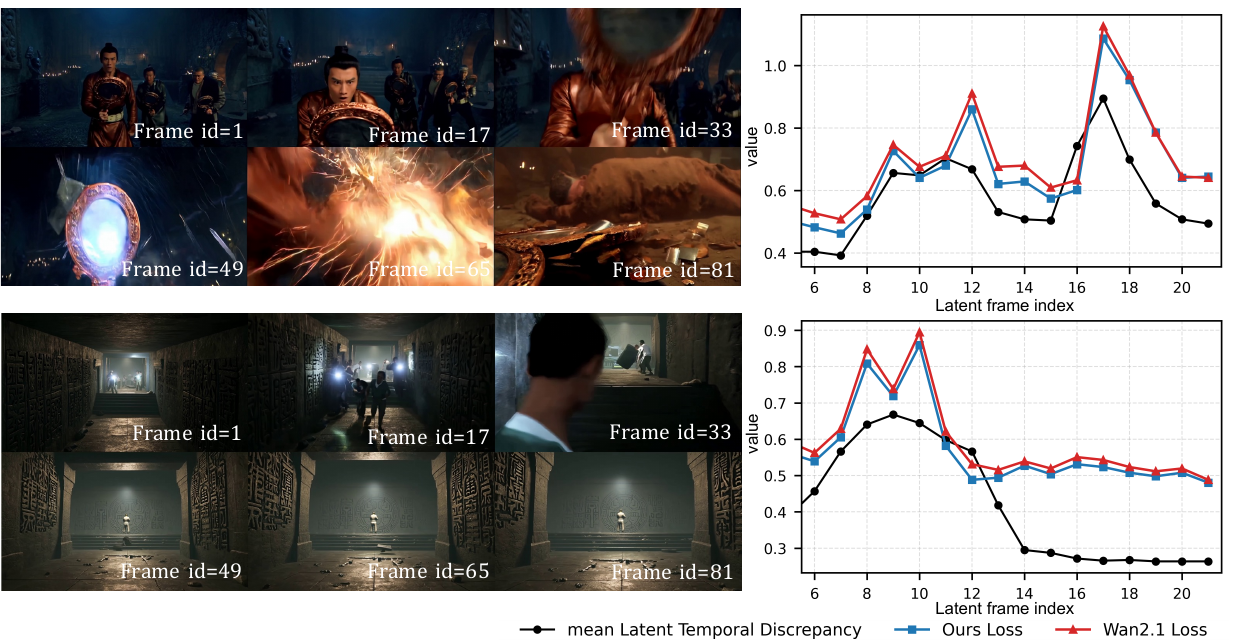}
    \captionsetup{aboveskip=2pt, belowskip=0pt}
    \caption{Line plot of mean Latent Temporal Discrepancy, Wan2.1 MSE loss, and Ours MSE Loss across latent frames.}
    \vspace{-3pt}
    \label{fig:method-s}
\end{figure}

We directly leverage the changes in latent features as the motion prior, as illustrated in Fig.\ref{fig:method}. Given a slide window size $\tau \in \mathbb{N}^+$, for each latent video frame $\mathbf{z}(f) \in \mathbb{R}^{H_l \times W_l \times C_l}$, where $f \in \{1, \ldots, F_l\}$ is the frame index, define the dynamic intensity as:

{\small
\begin{equation}
\begin{split}
    & D_f = \frac{1}{R_f - L_f} \sum_{i = L_f}^{R_f - 1} \| \mathbf{z}(i+1) - \mathbf{z}(i) \|, \\ 
    & L_f = \max\left(1,\; f - \left\lfloor \frac{\tau}{2} \right\rfloor \right), \quad
    R_f = \min\left(F_l,\; f + \left\lfloor \frac{\tau}{2} \right\rfloor \right).
\end{split}
\end{equation}}

To suppress extreme values in $D_f$ and enhance numerical stability, we apply a logarithmic transformation to the original dynamic magnitude $D_f$, obtaining the weighting factor $\omega$ for the loss function. The final loss function is:
{\small
\begin{equation}
    \mathcal{L} = \mathcal{L}_\mathrm{diffusion} + \mathrm{ln}(e+D) \times \mathcal{L}_\mathrm{diffusion},
\end{equation}}
\noindent
where $D$ denotes the spatiotemporal collection of $D_t$, and serves as a per voxel weighting tensor over the latent space with dimensions $F_l \times H_l \times W_l$.

\vspace{-3pt}
\section{Experiment}
\vspace{-3pt}
\subsection{Implementation Details}
\begin{figure}
    \centering
    \includegraphics[width=0.9\linewidth]{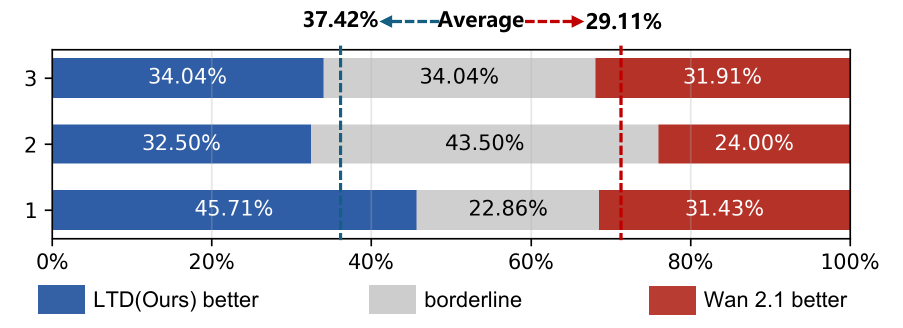}
    \captionsetup{aboveskip=2pt, belowskip=0pt}
\caption{Human Evaluation. The figure illustrates, for each participant, the preference ratios for Ours, Wan2.1~\cite{wan2.1}, and Indistinguishable cases.}
    \label{tab:placeholder}
\end{figure}

\begin{table*}
\caption{Ablation results evaluating the contribution of the LTD loss. Wan2.1~\cite{wan2.1} serves as the baseline, and adding LTD loss yields consistent improvements.}
\label{tb:ablation}
\centering
\resizebox{\linewidth}{!}{
\begin{tabular}{@{}l|cccccccc@{}}
\toprule
    \#       & \begin{tabular}[c]{@{}c@{}}Aesthetic \\ Quality(\%) $\uparrow$\end{tabular} 
             & \begin{tabular}[c]{@{}c@{}}Background \\ Consistency(\%) $\uparrow$\end{tabular} 
             & \begin{tabular}[c]{@{}c@{}}Dynamic \\ Degree(\%) $\uparrow$\end{tabular} 
             & \begin{tabular}[c]{@{}c@{}}Image \\ Quality(\%) $\uparrow$\end{tabular} 
             & \begin{tabular}[c]{@{}c@{}}Motion \\ Smoothness(\%) $\uparrow$\end{tabular} 
             & \begin{tabular}[c]{@{}c@{}}Subject \\ Consistency(\%) $\uparrow$\end{tabular} 
             & \begin{tabular}[c]{@{}c@{}}Temporal\\ Flickering(\%) $\uparrow$\end{tabular} 
             & \begin{tabular}[c]{@{}c@{}}Quality \\ Score(\%) $\uparrow$\end{tabular} \\ \midrule
Wan2.1 (Base)     & 64.41            & 96.26           & \textbf{70.83}   & 65.06             & 97.88      & 93.49        & 99.51       & 83.76  \\
w/o LTD Loss & 61.51            & 96.68           & 61.11            & 67.83             & 97.95      & \textbf{96.00}    & 99.40       & 83.52 \\
w/ LTD Loss  & \textbf{65.80}   & \textbf{97.16}  & 68.06            & \textbf{67.85}    & \textbf{98.39}   & 95.65   & \textbf{99.69} & \textbf{85.11} \\ \bottomrule
\end{tabular}
}
\end{table*}

\begin{figure*}[t!]
    \centering
    \includegraphics[width=\linewidth]{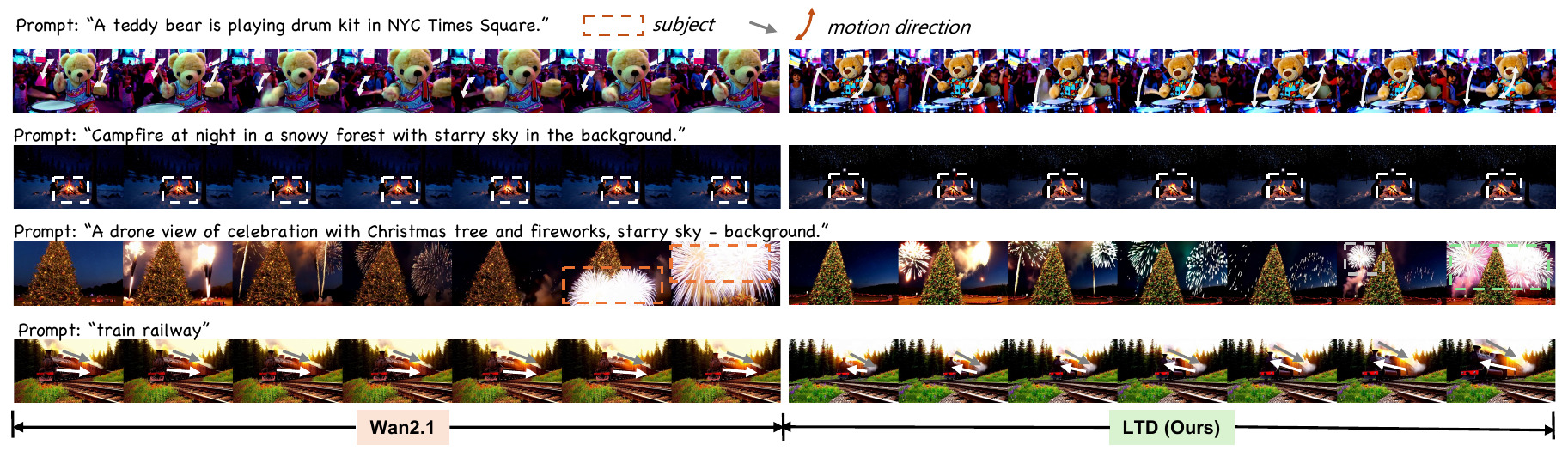}
    \captionsetup{aboveskip=2pt, belowskip=0pt}
    \caption{Qualitative comparison of generated videos. Wan2.1~\cite{wan2.1} may generate counterfactual motion directions where the subject violates physical plausibility, while our method yields more reasonable outcomes.}
    \label{fig:visual}
\end{figure*}

\noindent{\textbf{Model Training and Inference.}} We fine-tune our model on Wan2.1 \cite{wan2.1}, freezing the Variational Auto-Encoder (VAE) to preserve the pretrained latent space, while fully updating the parameters of the DiT blocks \cite{peebles2023scalable} during training. We use an internal lecensed dataset of 3,860 video-text pairs to fine-tune.
All videos are center-cropped spatially to $832\times480$ and temporally truncated to 81 frames during training to ensure consistent input dimensions.
The model is trained for 2 epochs on 8 H20 GPUs with a fixed learning rate of $2\times 10^{-5}$, slide windows size of 3, a batch size of 16, and 1000-step linear diffusion. At inference, we use DDIM \cite{song2020denoising} with 50 steps and a guidance scale of 7.5.

\vspace{-5pt}
\subsection{Comparison to Prior Works}
\vspace{-3pt}
We compare to 6 open-sourced T2V generation methods ($e.g.$, VideoCrafter~\cite{chen2024videocrafter2}, Wan2.1~\cite{wan2.1}, CogVideoX~\cite{yang2024cogvideox}, Hunyuan~\cite{kong2024hunyuanvideo}, OpenSora~\cite{opensora2} and OpenSora-Plan~\cite{lin2024opensora-plan}). We ensure fair comparisons by following each method’s pre-processing and post-processing pipeline to ensure the input is in the optimal aspect ratio and the generated videos are without any undesirable distortion. 

The results are summarized in Tab.~\ref{tb:sota}. We evaluate our method using both the general VBench-prompt~\cite{zheng2025vbench} and the motion-specific VMBench-prompt~\cite{ling2025vmbench}. The experiments show that our approach consistently surpasses prior methods on VBench, and achieves an 3.58\% improvement over Wan2.1 (Base) on VMBench. These results demonstrate that our method provides clear advantages, particularly in capturing motion dynamics.

\vspace{-5pt}
\subsection{Human Evaluation}
\vspace{-3pt}
We conduct a human evaluation to compare the video quality of Wan2.1 (Base)~\cite{wan2.1} and our method, covering 460 video pairs in total. Three annotators make their selections only after watching both videos in full, and we randomize the presentation order to avoid bias. We report a human preference score for each method, defined as the proportion of pairwise comparisons in which the video generated by that method is preferred by the annotators (via majority vote).

As shown in Fig.~\ref{tab:placeholder}, our method is preferred in 37.42\% of the comparisons, while Wan2.1 is preferred in 29.11\%, yielding a relative gain of 8.31\%. The remaining cases were judged as indistinguishable. These results indicate that human subjects consistently favor our method, which aligns well with the improvements observed on VBench~\cite{zheng2025vbench}. 
From the justification selections, the main reasons for our method to win are on the object motion, which is exactly the motivation of our method to improve text-driven motion learning.

\vspace{-6pt}
\subsection{Ablation Study}
\vspace{-3pt}
\noindent{\textbf{Latent temporal discrepancy Loss.}} To validate the effectiveness of the proposed LTD Loss, we fine-tune Wan2.1 with the same training data and identical settings, the only difference being the inclusion or exclusion of LTD loss. As shown in Tab.\ref{tb:ablation}, our method outperforms the Wan2.1 fine-tuned baseline across all metrics except for \textit{subject consistency}, where it achieves slightly lower scores. Notably, our approach excels in motion-related aspects, achieving significantly higher ratings in \textit{motion smoothness} and \textit{dynamic degree}, indicating superior temporal coherence and motion expressiveness. These results demonstrate the effectiveness of the proposed LTD Loss in enhancing motion realism and temporal consistency, while maintaining overall generation quality.

Moreover, as shown in Fig.~\ref{fig:method-s}, the loss variation follows the LTD trend. For video segments on the left with drastic motion changes, the corresponding peaks can be observed in the right-hand loss curve. Our loss reweighting strategy effectively reduces these peaks, leading to more stable optimization and improved learning of dynamic regions.

\vspace{-5pt}
\subsection{Visual Analysis}
\vspace{-3pt}
We present four representative visualization results for qualitative analysis as shown in Fig.\ref{fig:visual}. In the video generated with the prompt \textit{“A teddy bear is playing drum kit in NYC Times Square,”} Wan2.1 fails to animate coherent drumming motion, showing the bear stuck in a raised-stick pose, while our method produces a more natural and fluid motion. In the prompt \textit{“Campfire at night in a snowy forest with starry sky in the background,”} our method generates more natural and dynamic flames, while Wan2.1 produces nearly static fire. For the prompt \textit{“A drone view of celebration with Christmas tree and fireworks, starry sky – background,”} our model renders fireworks that appear abruptly and dissolve realistically, whereas Wan2.1 exhibits spatial-temporal inconsistencies, with fireworks and the Christmas tree misplaced in the distant background. In the \textit{“train railway”} case, Wan2.1 generates a train moving backward, accompanied by smoke flowing in the wrong direction, while our method produces motion that is physically plausible and temporally coherent.

\vspace{-3pt}
\section{Conclusion}
\vspace{-3pt}
This work shows that incorporating a motion-aware prior through Latent Temporal Discrepancy effectively alleviates the instability of static loss in video diffusion models. The proposed loss reweighting not only improves dynamic fidelity but also enhances temporal consistency, as validated on both VBench and VMBench. We believe this provides a practical step toward more reliable motion-aware text-to-video generation.
\clearpage

\vfill\pagebreak

\apptocmd{\thebibliography}{\small\setlength{\itemsep}{3pt}}{}{}
\bibliographystyle{IEEEbib}
\bibliography{strings,refs}

\end{document}